\documentclass[letterpaper, 10 pt, conference]{ieeeconf}  

\IEEEoverridecommandlockouts                           
\makeatletter
\let\NAT@parse\undefined
\makeatother
\usepackage{hyperref}
\hypersetup{colorlinks=true, linkcolor=red, citecolor=blue, urlcolor=magenta}

\usepackage[utf8]{inputenc}
\usepackage[T1]{fontenc}
\usepackage{graphicx}
\usepackage{subcaption}
\usepackage{amsmath}
\usepackage{amssymb}

\title{\LARGE \bf
Tightly-Coupled LiDAR-Visual SLAM Based on Geometric Features for Mobile Agents
}

\author{Ke Cao$^{1}$, Ruiping Liu$^{1}$, Ze Wang$^{3}$, Kunyu Peng$^{1}$, Jiaming Zhang$^{1}$, Junwei Zheng$^{1}$,\\Zhifeng Teng$^{1}$, Kailun Yang$^{2,*}$, and Rainer Stiefelhagen$^{1}$%
\thanks{This work was supported in part by the Ministry of Science, Research and the Arts of Baden-Württemberg (MWK) through the Cooperative Graduate School Accessibility through AI-based Assistive Technology (KATE) under Grant BW6-03, in part by the University of Excellence through the ``KIT Future Fields'' project, in part by the Helmholtz Association Initiative and Networking Fund on the HAICORE@KIT partition, and in part by Hangzhou SurImage Technology Company Ltd.}%
\thanks{$^{1}$The authors are with Karlsruhe Institute of Technology, Germany.}%
\thanks{$^{2}$The author is with Hunan University, China.}%
\thanks{$^{3}$The author is with Zhejiang University, China.}%
\thanks{$^{*}$Corresponding author: Kailun Yang (E-Mail: kailun.yang@hnu.edu.cn).}%
}

\begin{document}

\maketitle
\thispagestyle{empty}
\pagestyle{empty}

\begin{abstract}
The mobile robot relies on SLAM (Simultaneous Localization and Mapping) to provide autonomous navigation and task execution in complex and unknown environments. However, it is hard to develop a dedicated algorithm for mobile robots due to dynamic and challenging situations, such as poor lighting conditions and motion blur. To tackle this issue, we propose a tightly-coupled LiDAR-visual SLAM based on geometric features, which includes two sub-systems (LiDAR and monocular visual SLAM) and a fusion framework. The fusion framework associates the depth and semantics of the multi-modal geometric features to complement the visual line landmarks and to add direction optimization in Bundle Adjustment (BA). This further constrains visual odometry. On the other hand, the entire line segment detected by the visual subsystem overcomes the limitation of the LiDAR subsystem, which can only perform the local calculation for geometric features. It adjusts the direction of linear feature points and filters out outliers, leading to a higher accurate odometry system. Finally, we employ a module to detect the subsystem's operation, providing the LiDAR subsystem's output as a complementary trajectory to our system while visual subsystem tracking fails. The evaluation results on the public dataset M2DGR, gathered from ground robots across various indoor and outdoor scenarios, show that our system achieves more accurate and robust pose estimation compared to current state-of-the-art multi-modal methods.

\end{abstract}


\section{INTRODUCTION}
In the field of mobile robots and autonomous navigation agents, there has been a notable increase of interest among different studies~\cite{pandey2017mobile}.
They are employed for a variety of purposes, including independent living for the elderly~\cite{cortes2007assistive}, guiding and assisting shoppers in large retail spaces~\cite{orciuoli2015agent}, and facilitating delivery services in outdoor industrial or commercial areas~\cite{abbenseth2017cloud}. When a robot enters a new scene or encounters an environment with updated details, it carefully plans actions to explore unknown or uncertain areas. The goal is to gather valuable information about the new terrain and improve the accuracy of its reconstructed map. Thereby, the robot utilizes the measurements provided by its sensors to reconstruct the map and perform localization through SLAM. 

   \begin{figure}[thpb]
      \centering
      \parbox{0.46\textwidth}{
        \centering
        \includegraphics[width=\linewidth]{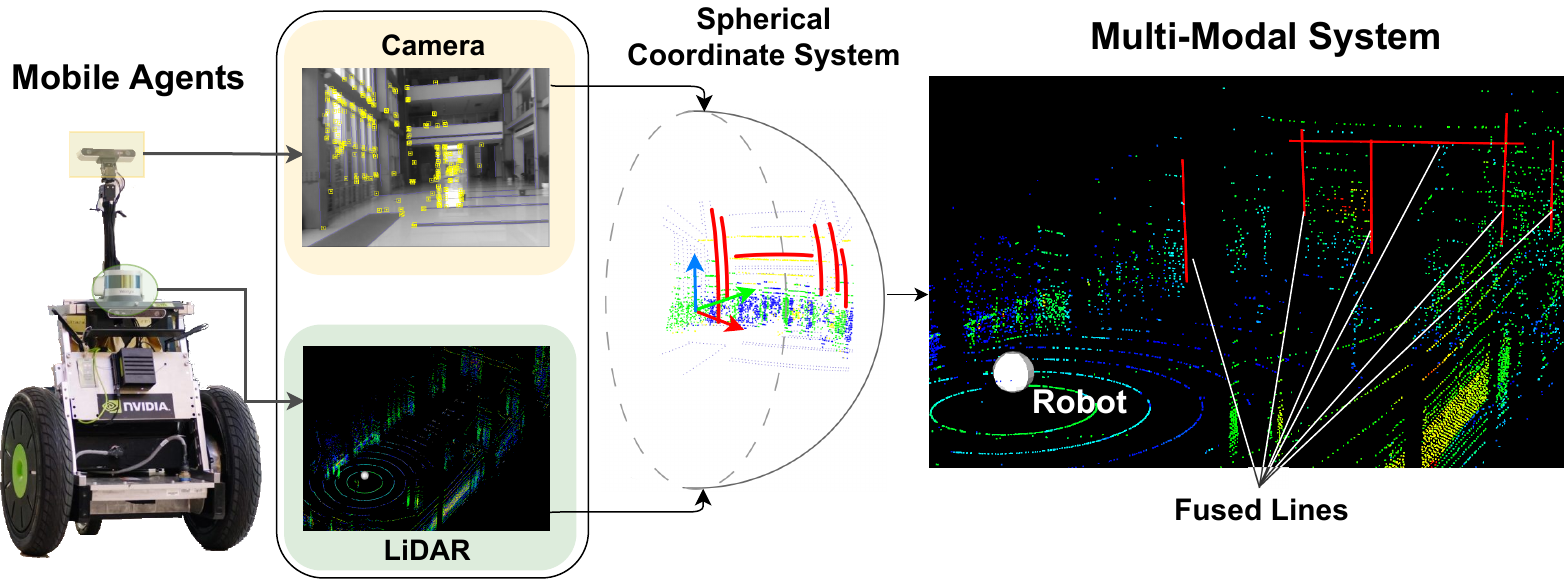}
        \caption{Our method projects the geometric features (extracted from the sensor camera and LiDAR data) onto the fusion frame (the spherical coordinate system). Through a series of optimizations, a more accurate trajectory and map are obtained with the reconstructed line.}
        \label{fig:overview_intro}
      }
      \vskip -3ex
   \end{figure}
   
Tracking failure in challenging environments has hindered the widespread adoption of mobile robots. While visual SLAM \cite{forster2014svo, mur2017orb} offers superior motion estimation, it tends to fail in extreme lighting conditions and outdoor environments. Similarly, LiDAR SLAM~\cite{shan2018lego, zhang2014loam} is highly reliable for tracking motion in large-scale scenes but faces challenges in degraded cases. Traditional SLAM methods rely heavily on a single source, which may be prone to failure in adverse scenarios or due to equipment issues. In recent years, significant approaches~\cite{chou2021efficient, shan2021lvi} have been developed to integrate measurements from multiple sensors to overcome the limitations of mono-sensor algorithms. However, some SLAM systems~\cite{chou2021efficient} combine camera, IMU, and LiDAR to recalibrate the extrinsic parameters which can be quite computationally expensive. Therefore, our research focuses on the development of a camera-LiDAR framework. The primary goal of this framework is to minimize sensor costs for mobile agents while conserving computational resources. Moreover, it aims to provide highly accurate mapping and localization in a variety of scenarios.

In practical situations, devices often oscillate when in motion. This leads to less accurate extrinsic calibration between the camera and the LiDAR. The fusion performance~\cite{graeter2018limo}, which relies solely on the association of feature points, is highly susceptible to the misalignment error caused by this oscillation. Consequently, inaccurate depth estimates are obtained, compromising the overall accuracy of the system.
Other SLAM methods~\cite{fang2020visual, lee2021plf} that depend on geometric features prove that lines are more reliable and stable. Therefore, our work is exclusively concentrated on geometric-level fusion to combine the monocular camera and LiDAR sensors.

In this paper, we propose a multi-modal SLAM system that tightly integrates parallel monocular visual SLAM and LiDAR SLAM subsystems. This design ensures that if one subsystem fails, the other subsystem continues to operate, providing a more robust system for robot navigation. Moreover, we utilize more stable linear and planar features to minimize the impact of inaccurate extrinsic calibration. In Figure~\ref{fig:overview_intro}, each subsystem utilizes perceptual data from the mobile robot's sensors (a camera and a LiDAR sensor) to extract geometric features. We align the features from both subsystems in terms of temporality, spatiality, and dimensionality through a unified reference fusion frame. After fusion, the reconstructed features contribute as new landmarks to the initial pose estimation in the visual subsystem. They are utilized as additional optimization parameters in the back end, where the endpoint and direction are used to constrain the visual odometry. In the LiDAR subsystem, the direction of the geometric features is adjusted by the detected line from the visual system, reducing the probability of outliers during registration.

We evaluate our approach on the M2DGR dataset \cite{yin2021m2dgr}, which consists of various indoor and outdoor environments commonly encountered in mobile robot applications. By comparing with our visual baselines Structure PLP-SLAM~\cite{shu2022structure}, LiDAR baseline MULLS~\cite{pan2021mulls}, and multi-modal algorithms LVI-SAM~\cite{shan2021lvi} and ORB-SLAM3~\cite{campos2021orb}, our SLAM method achieves superior accuracy and robustness in various challenging environments. It is particularly suitable for low-cost mobile robots in various applications.

To summarize, we present the following contributions:

\begin{itemize}
\item We build a fusion framework -- a spherical coordinate system -- to maintain spatial and temporal consistency. This framework integrates the geometric features from the visual subsystem and the LiDAR subsystem.

\item In the LiDAR subsystem, we optimize the linear directions to improve registration efficiency and accuracy, while increasing the probability of optimized points in its local map.

\item We reconstruct more lines for pose estimation in the visual subsystem. In the back end of the visual subsystem, we propose a new optimization term, \textit{i.e.}, line direction term, and more fusion lines as optimization parameters to constrain the trajectory.

\end{itemize}

\section{RELATED WORK}
   \begin{figure*}
      \centering
      \parbox{0.6\textwidth}{
        \centering
        \includegraphics[width=\linewidth]{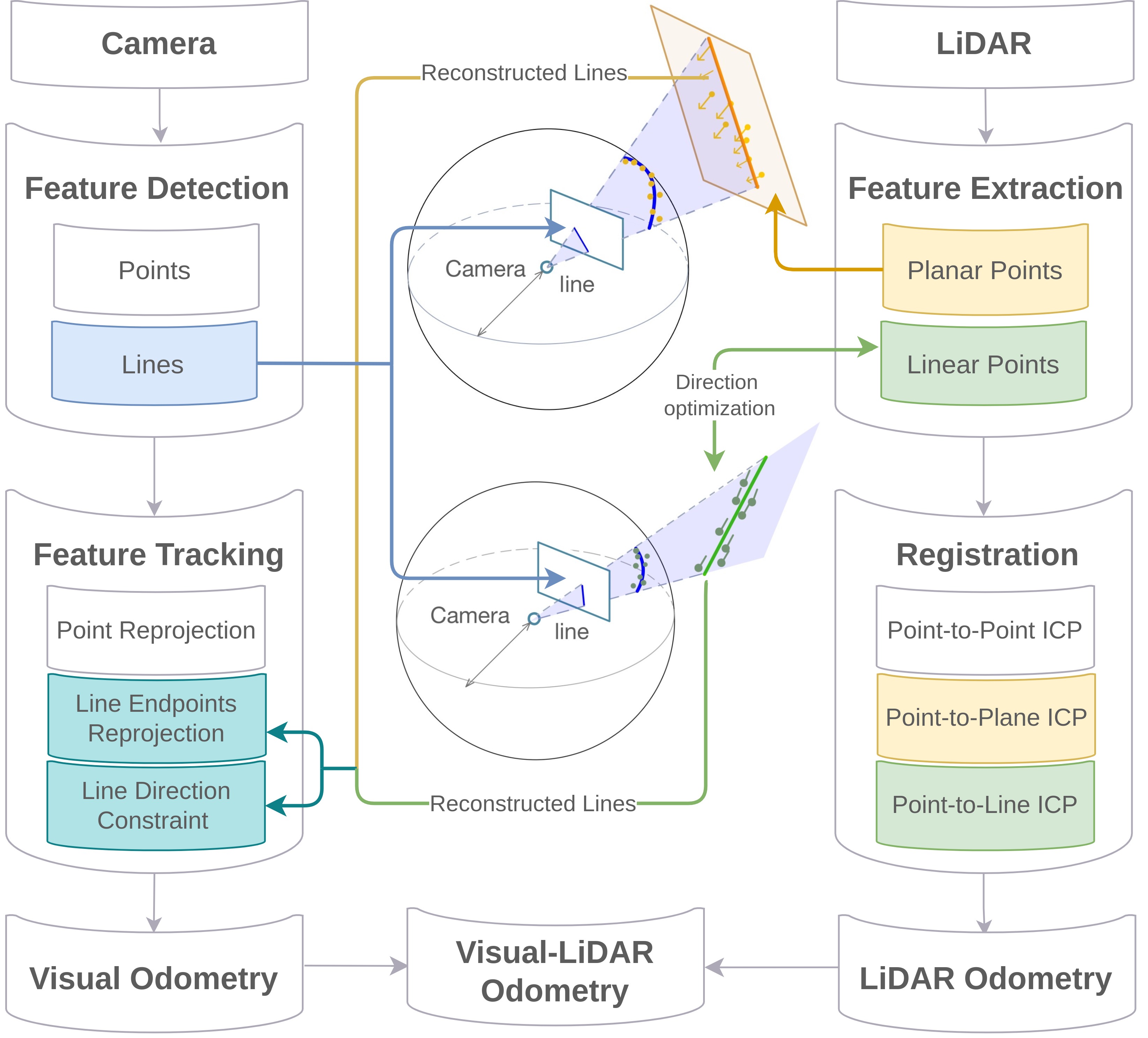}
        
      }
      \caption{Overview of our multi-modal SLAM system. It consists of a fusion framework (middle section) and dual systems - the visual SLAM subsystem (left flow) and the LiDAR SLAM subsystem (right flow). The geometric features are extracted by subsystems and then passed to the fusion framework to perform the fusion computations. The generated lines are returned to the subsystems to optimize their odometry and map.}
      \label{fig:overview}
      \vskip -3ex
   \end{figure*}
   
\noindent\textbf{Multi-modal SLAM.}
Previous multi-modal works can be classified by the type of sensors.
Visual SLAM~\cite{campos2021orb} optimizes the reprojection error and temporal error by adding other information from IMU and correcting the scales.
Filter-based algorithms~\cite{chen2018review, geneva2020openvins} use inertial measurements for state propagation and update the visual data to improve accuracy.

In the field of high-cost and high-complexity visual-LiDAR-inertial SLAM systems, the optimization-based algorithm LVI-SAM~\cite{shan2021lvi}, consisting of two subsystems, builds on a factor graph and accomplishes tightly-coupled odometry.
TVL-SLAM~\cite{chou2021efficient} proposes a novel large-scale bundle adjustment optimization, which compresses the LiDAR and visual residuals to achieve real-time performance in the degraded environment.
Extended Kalman Filter (EKF) based algorithms~\cite{lin2021r} contribute to enhancing robustness in environments with weak texture or even no features.

The visual-LiDAR SLAM systems~\cite{seo2019tight, shin2020dvl} provide a low-cost, low-compute, high-precision method for mapping and calculating odometry.
DEMO~\cite{zhang2014real} is the first to associate LiDAR depth measurements with Harris corner features.
However, due to the lack of loop closure, the accumulated residuals cannot be optimized and eliminated.
Liang~\textit{et al.}~\cite{liang2016visual} address this issue by proposing a scan-matching method with a visual loop detection scheme using ORB features~\cite{rublee2011orb} and a Bag-of-Words model. 

However, the extracted point-only depth as prior factors has a flaw: The multi-sensor aligned error caused by mechanical changes cannot be mitigated without an automatic extrinsic calibration procedure.
LIMO~\cite{graeter2018limo} projects the point cloud onto the image and estimates the point depth by a subset of points within a fixed rectangular region around it. To minimize the number of outliers, a filtering mechanism was introduced. This mechanism limits the depth estimation to subsets within planes.
Huang~\textit{et al.}~\cite{huang2020LiDAR} explore prior structural information for point-line Bundle Adjustment (BA) and create a novel scale correction scheme.
Although it reduces the system's sensitivity to noise and motion blur, LiDAR-enhanced visual odometry becomes more challenging in underexposed scenarios.
Differing from existing works, our approach is unique in that we incorporate LiDAR-enhanced visual odometry while also developing visual-enhanced LiDAR odometry. This method is designed to be more robust in undesirable lighting conditions.

\noindent\textbf{Line-based SLAM.}
Line-based SLAM performs robustly in man-made scenes, especially when point features are sparse or unevenly distributed in images.
Its map exhibits remarkable richness, comprising a diverse range of geometric elements. Therefore, feature-based and direct visual SLAM systems incorporate geometric features to enhance localization accuracy and robustness in scenes with weak textures and lighting variations.
Visual SLAM~\cite{shu2022structure,ram2021rp}, incorporating plane features, increases the computational complexity of feature extraction and matching.

Feature-based methods~\cite{gee2006real, gomez2019pl} use traditional line detection algorithms like LSD~\cite{engel2014lsd} and perform descriptor-based tracking by minimizing the line projection.
Since the line has only four Degrees of Freedom (DoF), the two-endpoint representation line introduces six parameters, resulting in overparameterization.
Bartoli and Adrien~\cite{bartoli2005structure} propose an orthogonal representation that uses a three-DoF rotation matrix and a one-DoF rotation matrix to update line parameters during optimization.

Lines determined by two-view triangulation often occur during subsequent tracking, which wastes computational resources for detecting, describing, and matching line features.
Problems like line degeneracy occur more frequently when it is close to the epipole.
Lee~\textit{et al.}~\cite{lee2019elaborate} suggests reconstructing the line segments with multiple views instead of just two.
Furthermore, several algorithms~\cite{lim2021avoiding, yang2019visual} investigate different scenarios of degenerate motion.
They modify LSD and set additional parameters to extract discernable long-line features to prevent this degeneration.

Many visual SLAMs~\cite{georgis2022vp, wang2021vanishing} extract vanishing points to add constraints associated with parallel 3D lines to optimize robot pose estimation.
In addition, the PLF-VINS algorithm~\cite{lee2021plf} also integrates point-line coupling residuals based on the similarity of corner points and line features.
This integration ensures accurate depth estimation.

Line-based visual SLAM is commonly used in man-made indoor scenes. Unlike existing works, our approach stands out by combining the accurate geometric features of a LiDAR point cloud with the overall structural information from an image. This results in more effective utilization of line features in a wider range of scenarios.

\section{METHODOLOGY}
\subsection{System Overview}

The flowchart of the proposed multi-modal system is shown in Figure~\ref{fig:overview}, which contains the hardware and software frameworks.
The hardware framework consists of a pinhole camera and a solid-state multi-beam LiDAR sensor.
The popular and consumer-grade RGB camera serves as an efficient sensor for the robot agent due to its high performance. The LiDAR sensor captures a comprehensive $360${\textdegree} horizontal surround view of the environment, providing stable depth information even in challenging real-world scenarios. Its long-range capabilities make it well-suited for outdoor usage.

The software framework, a tightly-coupled multi-modal SLAM system, comprises four primary modules: a fusion framework, a detection module, and two subsystems, \textit{i.e.}, the LiDAR SLAM system and the monocular visual SLAM system.
These subsystems perform odometry estimation and mapping tasks independently while exchanging geometric information.

After receiving sensor data, each subsystem proceeds to extract geometric features. The visual system utilizes ORB~\cite{rublee2011orb} and LSD~\cite{engel2014lsd} algorithms to detect points and lines in images. On the other hand, the LiDAR system employs the PCA algorithm~\cite{hackel2016fast} to extract point, linear, and planar features. To overcome the limitations imposed by individual sensor types, the fusion framework 
combines the geometric features and projects them into spherical coordinates, creating a unified representation that incorporates both temporal and spatial dimensions. By performing a K-D tree search, we establish a definitive correlation between the detected lines and the depths of geometric features in the point cloud. This correlation allows us to reconstruct depth measurements for line segments that cannot be accurately triangulated by the visual subsystem alone.

   \begin{figure}[thpb]
      \centering
      \parbox{0.44\textwidth}{
        \centering
        \includegraphics[width=\linewidth]{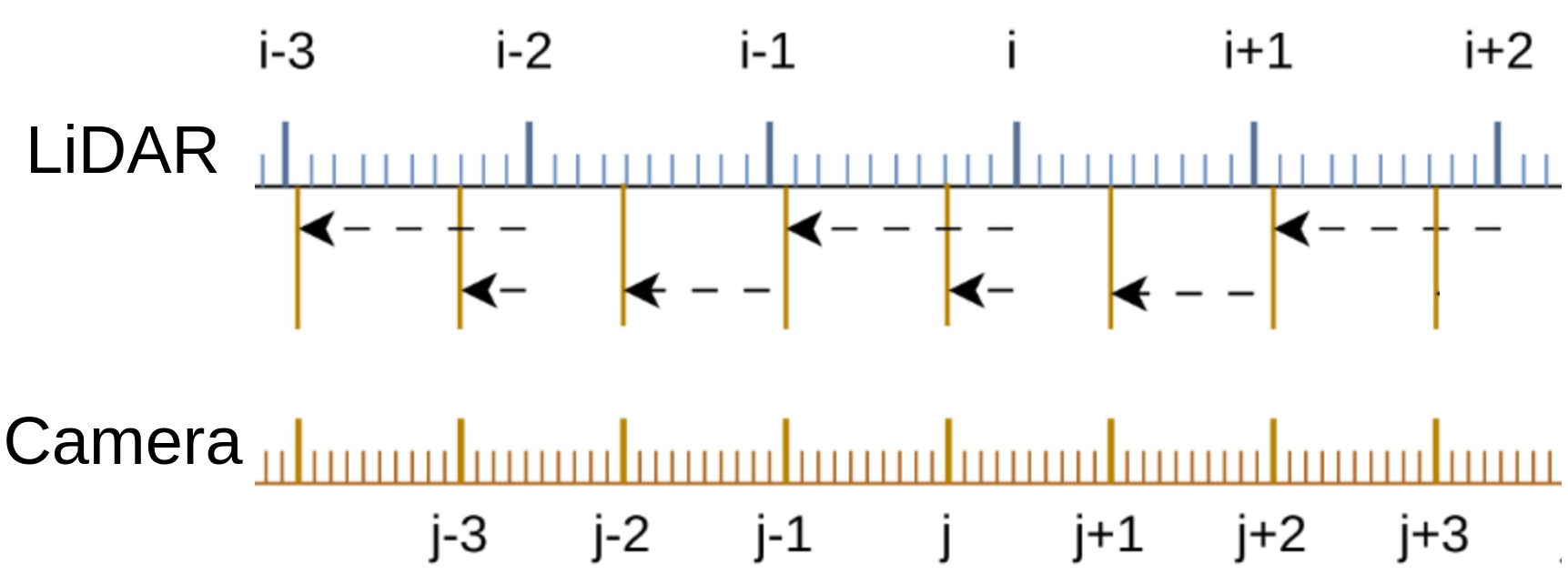}
      }
      \caption{LiDAR sensor data is aligned to the camera timestamp for temporal consistency.}
      \label{fig:timestamp}
      \vskip -3ex
   \end{figure}
   
Once reconstructed by the fusion framework, the lines serve as new features with precise depth and are fed back to the subsystems for motion estimation and optimization. In the LiDAR subsystem, the linear directions along the same line can be inconsistent due to sparse point clouds. 
The fusion framework determines their average direction, which is used to adjust the directions of the linear features in the LiDAR subsystem, reducing the occurrence of outliers. When both the fusion framework's data association and the visual subsystem's triangulation provide the direction of the same line segment, the visual subsystem introduces a new optimization term called the ``line direction residual''. This term evaluates the angular difference between the two directions. By iteratively minimizing this residual, the system effectively constrains the direction of the line landmarks.

Our system relies on the visual subsystem's output to determine its odometry, which exhibits higher precision in scenes with abundant texture.
However, given the increased likelihood of visual SLAM tracking failure, we incorporate a detection module to verify the functioning of the visual subsystem.
When brightness decreases or the sensor shakes strongly, the visual system might not capture enough feature points. In response, our system switches to the more reliable LiDAR subsystem to perform pose estimation.
This crucial capability ensures enhanced robustness, enabling the system to maintain accurate localization and mapping even in challenging scenarios.

\subsection{Fusion Framework}
\noindent\textbf{Preprocessing - temporal and spatial alignment.} The integration of sparse geometric features from sensors with varying positions and frequencies inherently poses challenges within the unaligned fusion framework. We employ data accumulation and temporal and spatial consistency, enhancing the reliability of our approach effectively.

By accumulating multi-frame point clouds, our system achieves a higher level of detail in the point cloud, comparable to that of the camera image.  This increased resolution allows easier matching of geometric features between the point cloud and the camera image.

\begin{figure*}
  \centering
  \begin{subfigure}{0.135\linewidth}
    \centering
    \includegraphics[width=\linewidth]{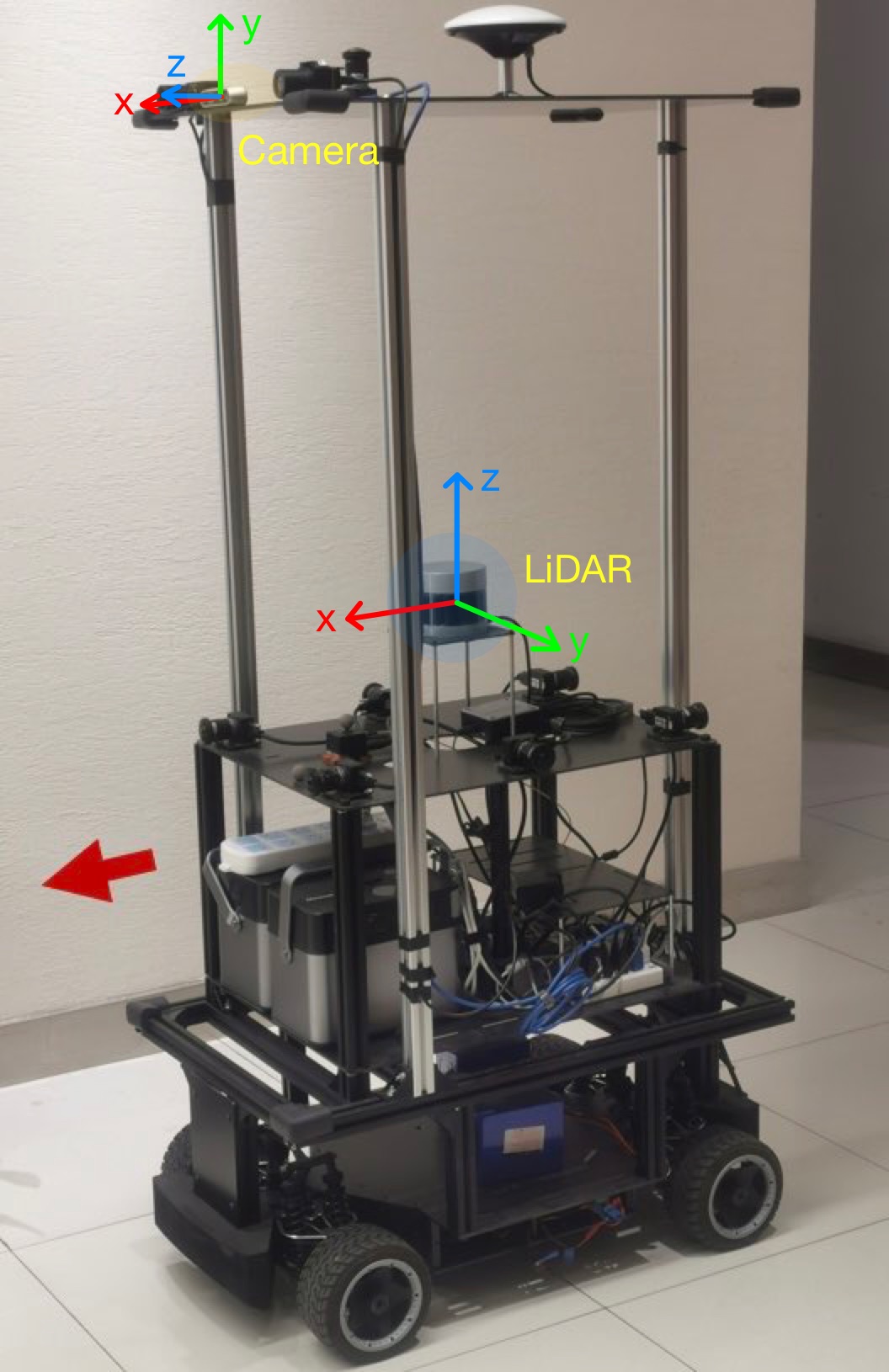}
    \caption{Ground robot}
    \label{fig:ground_robot}
  \end{subfigure}
  \hfill
  \begin{subfigure}{0.28\linewidth}
    \centering
    \includegraphics[width=\linewidth]{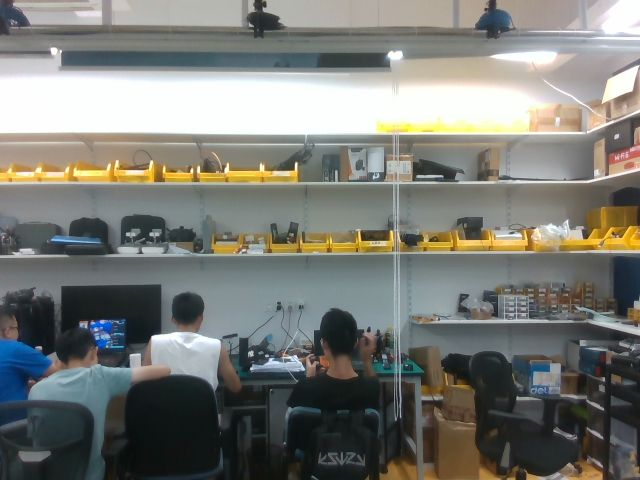}
    \caption{Indoor scene: room}
    \label{fig:room}
  \end{subfigure}
  \begin{subfigure}{0.28\linewidth}
    \centering
    \includegraphics[width=\linewidth]{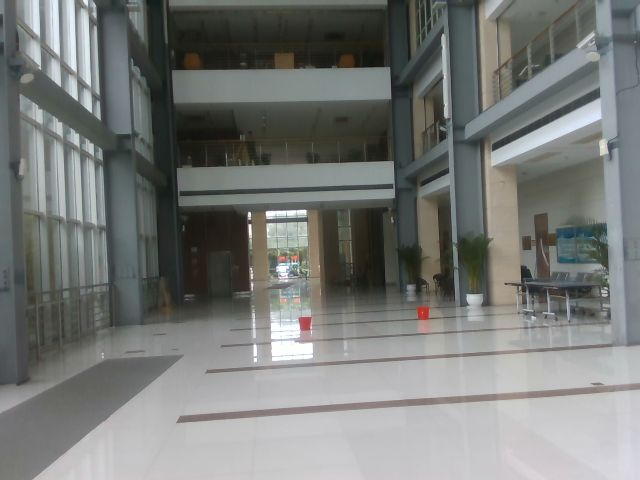}
    \caption{Indoor scene: hall}
    \label{fig:hall}
  \end{subfigure}
  \begin{subfigure}{0.28\linewidth}
    \centering
    \includegraphics[width=\linewidth]{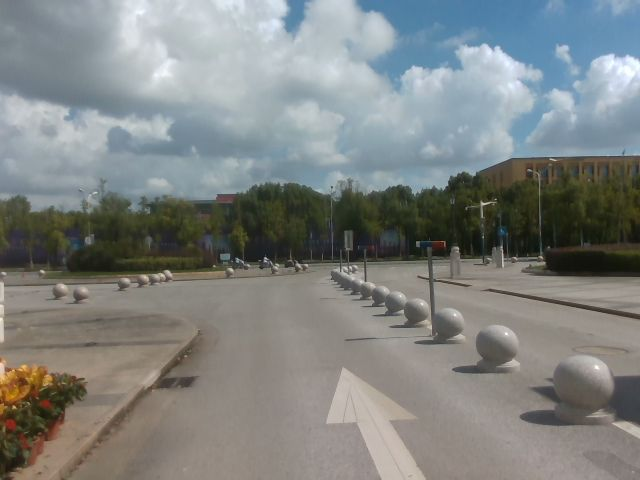}
    \caption{Outdoor scene: gate}
    \label{fig:outdoor}
  \end{subfigure}

  \caption{An overview of the mobile agent and scenes in the M2DGR dataset~\cite{yin2021m2dgr} to accomplish the dual-system SLAM.}
  \label{fig:dataset}
  \vskip -3ex
\end{figure*}

Figure \ref{fig:ground_robot} demonstrates that the camera and LiDAR sensors are mounted at different angles and positions. To integrate data from both sensors into a unified coordinate system, we leverage extrinsic calibration. This calibration provides the necessary information about the relative transformation between the camera and LiDAR, enabling accurate alignment of data acquired from both sensors.

To maintain temporal consistency, it is crucial to synchronize data between LiDAR and the camera. We define the camera coordinate system and timestamp as the reference. As illustrated in Figure~\ref{fig:timestamp}, we align the LiDAR point cloud within the time interval from frame $j$ to $j+1$ to the timestamp $t_{j+1}$. During each sweep, we calculate the time difference from each point to the end of the sweep period $t_{j+1}$. Assuming a constant angular and linear velocity over this short period and that the initial pose remains the same as the previous frame, we employ linear interpolation to compute the relative pose for each point and project it onto the timestamp $t_{j+1}$. This process achieves temporal consistency within our system.

\begin{figure}
      \centering
      \begin{subfigure}{0.43\linewidth}
        \centering
        \includegraphics[width=\linewidth]{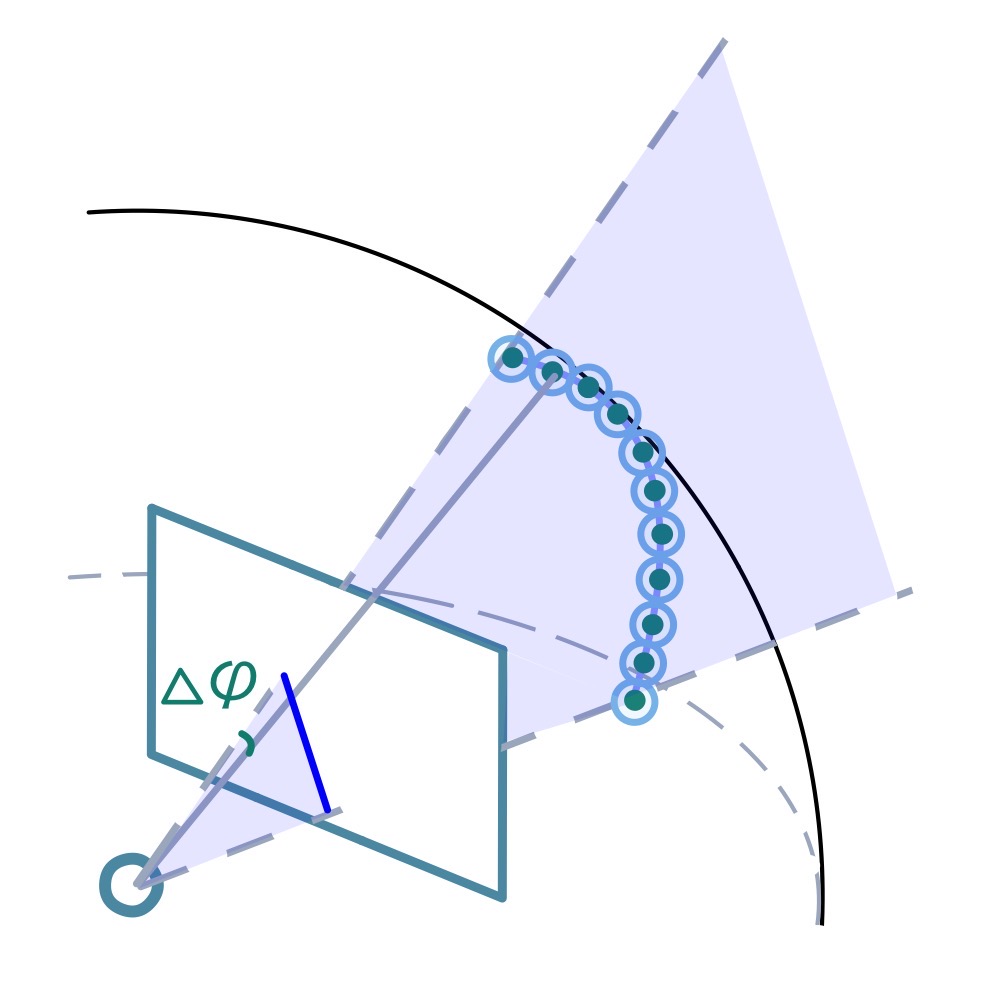}
        \caption{Decomposition of the arc into points}
        \label{fig:arc}
      \end{subfigure}
      \hfill
      \begin{subfigure}{0.43\linewidth}
        \centering
        \includegraphics[width=\linewidth]{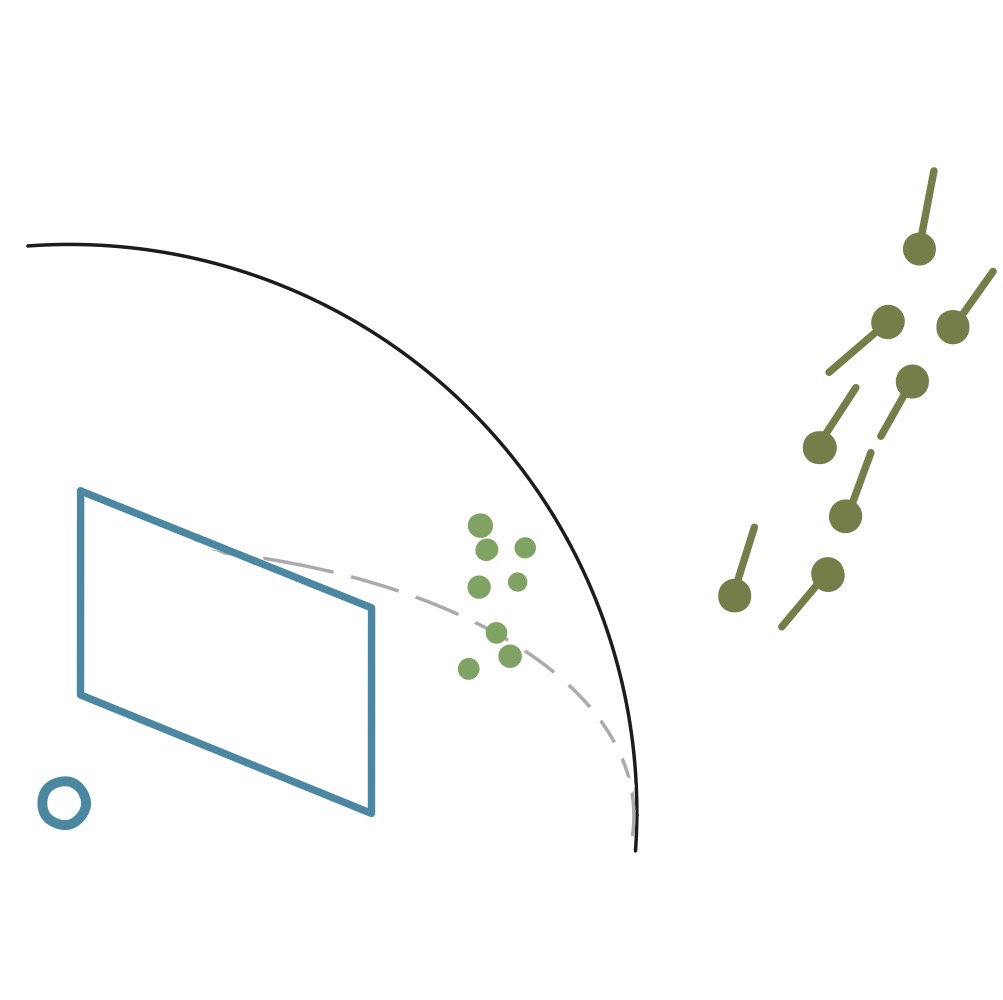}
        \caption{Direction of LiDAR linear points before optimization}
        \label{fig:lidar_bad_direction}
      \end{subfigure}
    
      \begin{subfigure}{0.43\linewidth}
        \centering
        \includegraphics[width=\linewidth]{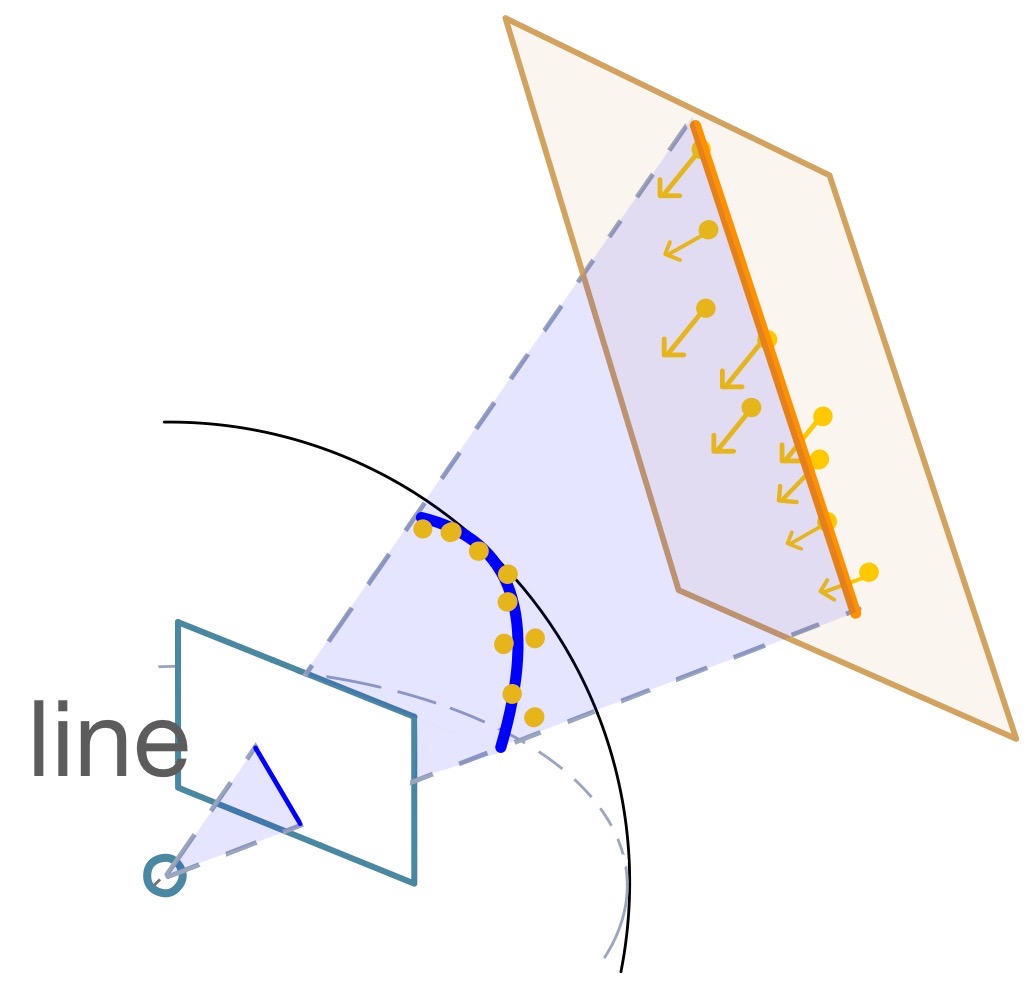}
        \caption{Reconstruction line based on LiDAR planar features}
        \label{fig:linear_reconst}
      \end{subfigure}
      \hfill
      \begin{subfigure}{0.43\linewidth}
        \centering
        \includegraphics[width=\linewidth]{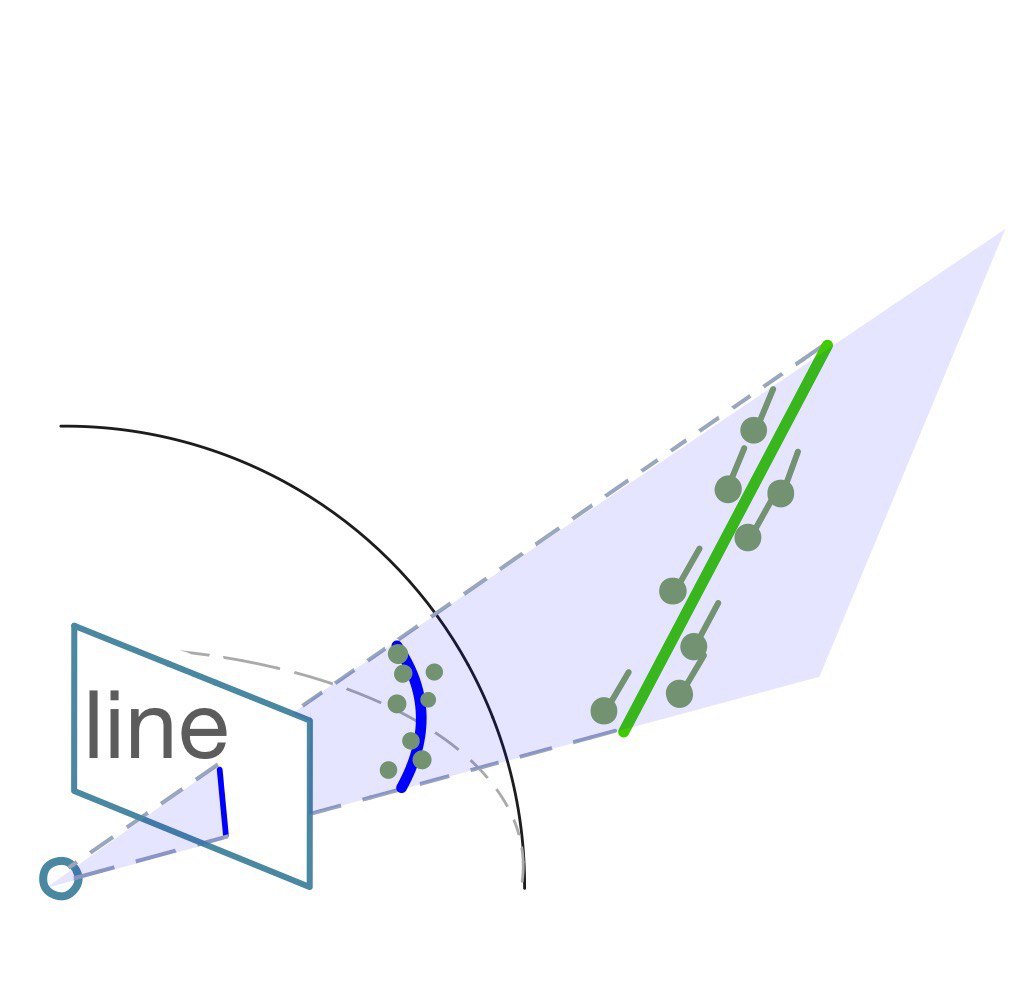}
        \caption{Reconstruction line based on LiDAR linear features}
        \label{fig:planar_reconst}
      \end{subfigure}
    
      \caption{Processing in the fusion framework.}
      \label{fig:four_images}
      \vskip -3ex
    \end{figure}

\noindent\textbf{Depth and direction association of geometric features.} After achieving spatial and temporal alignment, the geometric features from both subsystems are projected onto a spherical coordinate system centered at the camera. This projection is illustrated in Figure~\ref{fig:arc}, where a 2D visual line segment is projected onto the sphere as an arc. To perform the subsequent neighborhood search, we need to discretize the arcs and ensure a consistent density. To achieve this, we uniformly sample the arc at regular intervals $\Delta \varphi$ by incrementing the angle and determining the corresponding positions of the sampled points.
This sampling process results in a set of points along the arc with uniform spacing.
To process the LiDAR information, we first project the dense geometric point cloud onto a sphere. We then store all LiDAR geometric feature points in a 3D KD-tree~\cite{mark2008computational} and search for the closest geometric neighbor of each point along the visual arc. This iterative process continues until all points along the arc are successfully matched with their corresponding LiDAR feature points. This robust relationship between the visual arc and the LiDAR feature points allows for accurate line reconstruction. 

In the visual frame, we utilize the Line Segment Detector (LSD) algorithm to detect line features by identifying pixel regions with significant gradient changes. The first type comprises lines present on smooth surfaces, \textit{e.g.}, door or window frames. These features are extracted as planar features within the LiDAR point cloud. The second type encompasses linear features, including boundaries between different surfaces or well-defined linear elements, \textit{e.g.}, wall-floor boundaries, or trees. These characteristics are identified as linear features within the LiDAR point cloud.

Therefore, we divide the reconstruction lines, \textit{i.e.}, fusion lines, into two cases. When the majority of the surrounding point cloud comprises planar features, we utilize a plane-to-plane intersection approach to determine the fusion line, as depicted in Figure~\ref{fig:linear_reconst}. 
Firstly, we reconstruct a virtual LiDAR surface by utilizing appropriate planar feature points and their normal vectors. Subsequently, we project the arc into three-dimensional space, obtaining a projected plane. This plane is then intersected with the LiDAR plane to construct the fusion line.

Alternatively, if the visual arc, shown in Figure~\ref{fig:planar_reconst}, is surrounded by multiple linear features, the points $L_s$ are considered as part of a single line.
We identify clusters with similar directions and compute the average line direction, denoted as $L_{best}$.
With a point and the direction on a line, we determine the line segment based on these linear points.

The fusion frame establishes a spherical coordination system to effectively integrate features from multiple modalities. Under spatiotemporal consistency, the fusion process generates additional lines that are then fed back to the respective subsystems for more accurate localization and mapping.

\subsection{Visual Subsystem}
Our visual subsystem, a derivative of Structure PLP~\cite{shu2022structure}, performs visual SLAM by tracking feature points and feature lines. It comprises three distinct modules, \textit{i.e.}, the line extraction, the line reconstruction, and the joint optimization. Each module leverages information from the visual subsystem and fusion framework to modify and filter out inaccurate data, and assist in pose optimization. 

\noindent\textbf{Line detection.} During the feature extraction, we utilize the LSD~\cite{engel2014lsd} algorithm to detect line segments. To enhance the performance of LSD, we refine hidden parameters and implement length rejection, following the methodology of Structure PLP~\cite{shu2022structure}. These modifications contribute to improving the computational efficiency and accuracy of our visual system during feature extraction.

\noindent\textbf{Line reconstruction with triangulation} To reconstruct the detected lines in 3D space, we utilize two methods. The first method involves triangulating the matched line segments across two frames. By using the observed endpoints ${x_s, x_e}$ of a line segment $z$, we calculate the line vector ${l} = {x}_s \times {x}_e$. Using the camera projection matrix $P\in \mathbb{R}^{3 \times 4}$, which only contains the intrinsic camera parameters, we project the observed 2D line onto the camera coordinate system. This projection results in the projection plane $\pi_i = {l}_i^{\top} \mathrm{P}_i$.

   \begin{figure}[thpb]
      \centering
      \parbox{0.46\textwidth}{
        \centering
        \includegraphics[width=\linewidth]{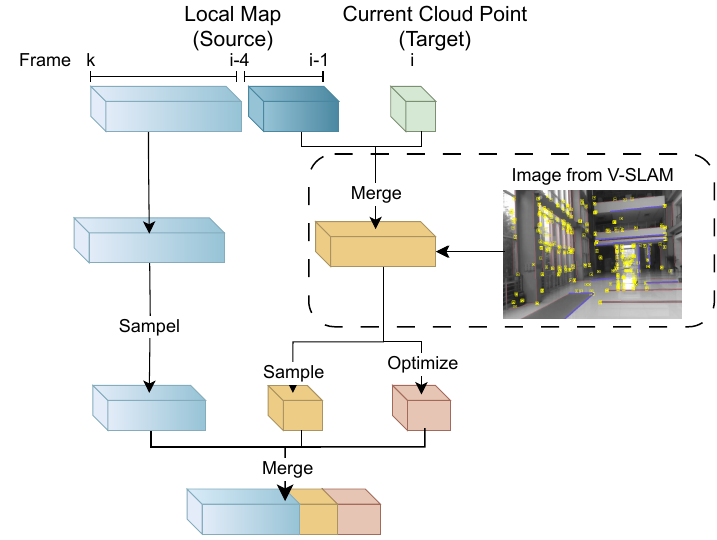}
        \caption{An overview regarding the local map update for ICP. The point cloud of the last four frames and the current frame participate in the fusion process with the visual subsystem. The optimized point cloud is kept or sampled less. The unoptimized point clouds are downsampled. Finally, all point clouds are merged to update the local map.}
        \label{fig:local_map}
      }
      \vskip -3ex
   \end{figure}
   
The intersection of the two projection planes $\pi_1$ and $\pi_2$ forms a 3D line in world coordinates. We represent this line using the Plücker coordinate system, denoted as $\mathbf{L}=\left(\mathbf{m}^{\top}, \mathbf{d}^{\top}\right)^{\top}$. To obtain its dual Plücker matrix $L^*_w$, the following calculation is performed:
\begin{equation}
\mathrm{L^*_w}=\pi_{1}\pi_{2}^{\mathrm{T}}-\pi_{2}\pi_{1}^{\mathrm{T}}\in\mathbb{R}^{4\times4}
\end{equation}

If the line cannot be successfully triangulated but is provided by the fusion framework, we need to convert this line segment from the Cartesian coordinate system to the Plücker coordinate system. 

Given two endpoints of the 3D lines $\overline{\mathbf{A}}$  and $\overline{\mathbf{B}}$, it is deduced that $\mathbf{L}=\left(\mathbf{m}^{\top}, \mathbf{d}^{\top}\right)^{\top}$ with
\begin{equation}
\left\{\begin{array}{l}
\mathbf{m}=\overline{\mathbf{A}} \times \overline{\mathbf{B}} \\
\mathbf{d}=\overline{\mathbf{B}}-\overline{\mathbf{A}}
\end{array}\right.
\end{equation}

\noindent\textbf{Endpoints trimming.} Line landmarks may have errors due to triangulation using only two frames. Similarly, line reconstruction in fusion frames can suffer from inaccuracies, particularly due to potential imperfections in the spatiotemporal alignment of two-sensor data. Therefore, we refine the endpoints with the method proposed in \cite{lee2019elaborate, zhang2015building}. 

In Structure PLP, endpoint trimming is utilized in local bundle adjustment (LBA). After this trimming, a depth check is performed with a threshold ratio of $0.1$ to filter out outliers based on the median depth change ratio of the scene. Considering the higher confidence in lines from the fusion framework, the outlier rejection threshold is set to $0.2$ for fusion line segments. 

\noindent\textbf{Endpoints re-projection and direction residual of line.} Once the observed line segment is detected, we calculate the distances between the endpoints $x_s$ and $x_e$ of the observed line segment $z$ and the projected line segment $l'$. The residual of the line measurement model is defined as the following reprojection error:
\begin{equation}
e_{dis}=d(z,l')=\begin{bmatrix}
\frac{\mathbf x_s^\top \mathbf l'}{\sqrt{l_1^{2}+ l_2^{2}} }   & \frac{\mathbf x_e^\top \mathbf l'}{\sqrt{l_1^{2}+ l_2^{2}} }
\end{bmatrix}^\top
\end{equation}

Many visual SLAM systems that use line features focus primarily on optimizing endpoints while ignoring their directions. Due to the considerable length of fusion line segments, their direction tends to be reliable. We introduce a new type of optimization, called ``line direction optimization'', which imposes additional constraints on the direction of line landmarks.

In our method, we compare the direction $\mathbf d_{fu}$ of the fusion line with the reconstructed line direction $\mathbf d_{cam}$ obtained from the visual subsystem. We unitize the two direction vectors and introduce an additional optimization term to measure the angle between these unit vectors:
\begin{equation}
\begin{aligned}
e_{dir}&=1-\cos (\mathbf d_{cam}, \mathbf d_{fu}) \\
& = 1 -
\frac{\mathbf d_{cam}^\top \mathbf d_{fu}}{\sqrt{d_{cam1}^{2}+ d_{cam2}^{2}} \sqrt{d_{fu1}^{2}+ d_{fu2}^{2}} }
\end{aligned}
\end{equation}

Considering the increasing gradient of this error term with larger errors, we incorporate a robust cost function. This effectively mitigates the influence of outliers, improving the overall stability of the optimization process. 
Moreover, since the angle measurement is independent of the image pyramid level, there is no need to introduce an information matrix.

\subsection{LiDAR Subsystem}
Following the LiDAR SLAM MULLS~\cite{pan2021mulls}, we propose modifications to optimize our LiDAR subsystem. 

\noindent\textbf{Modification of the linear direction.} The geometric characteristics of each point are analyzed using the Principal Component Analysis (PCA) algorithm~\cite{hackel2016fast, weinmann2013feature}. PCA determines the local neighborhood of a point and computes the eigenvalues $\lambda_{1} \geq \lambda_{2} \geq \lambda_{3} \geq 0$ and their corresponding eigenvectors $\mathbf{e}_{1}, \mathbf{e}_{2}, \mathbf{e}_{3}$. Following these methods~\cite{hackel2016fast, weinmann2013feature}, we classify points as either linear or planar based on their one-dimensional (1D) linear structure $L_{\lambda}=\frac{\lambda_{1}-\lambda_{2}}{\lambda_{1}}$, two-dimensional (2D) planar structure $P_{\lambda}=\frac{\lambda_{2}-\lambda_{3}}{\lambda_{1}}$, and curvature measurements $C_{\lambda}=\frac{\lambda_{3}}{\lambda_{1}+\lambda_{2}+\lambda_{3}}$. 
The geometric features are categorized into five types: vertex, pillar, beam, facade, and roof. 

In the PCA, the size of the local neighborhood is typically limited. This limitation poses challenges in larger scenes where the structural analysis may not be fully exploited. As depicted in Figure~\ref{fig:lidar_bad_direction}, the calculated line directions along the same line exhibit slight variations. These errors have a negative impact on the point-to-line alignment within the Iterative Closest Point (ICP) process. So we leverage detected complete lines in our visual subsystem to gain a more comprehensive understanding of the line structures in the environment. Following the fusion process, we determine which LiDAR linear points ($L_s$) belong to the same line with a consistent line direction. This association allows us to group related linear points together. Then, in the fusion framework, we calculate the mean direction ($\mathbf d_{v\_mean}$) of these lines and update the direction of all linear points by assigning them the calculated mean direction ($\mathbf d_{v\_mean}$).

\noindent\textbf{Improved storage of fusion-enhanced point cloud.} After performing the Iterative Closest Point (ICP) algorithm~\cite{besl1992method}, the target cloud must be updated for the next frame. MULLS merges and downsamples multiple historical frames into the local map. During this process, it is essential to synchronize the point clouds from different timestamps to the current timestamp $i$. This synchronization ensures that the point clouds are correctly aligned and accurately represent the environment. Although the denser local map provides a more comprehensive representation of the environment structure, merging too many historical frames means that the local map has to perform multiple alignments with the estimated pose, which leads to error accumulation. 

To strike a balance between preserving important geometric features and minimizing the accumulation of historical frames, a series of steps are employed in our method.
As shown in Figure~\ref{fig:local_map}, we combine the point cloud from the last four frames with the current frame. These point clouds are then projected onto a sphere to optimize the directions of their linear features.
The optimized feature points increase the convergence speed of the ICP and improve the accuracy of the estimated values. Once the point cloud is optimized, we proceed to sample and merge them. To control the density of the local map, we introduce a threshold parameter $\alpha$ that represents the maximum rate of optimized features allowed in the local map. If the current rate of optimized points does not exceed $\alpha$, no sampling is performed.
If the rate surpasses the threshold $\alpha$, we downsample the point cloud to reduce the rate until it falls below the specified threshold.
Finally, the local map is updated by merging all the point clouds.

\section{EXPERIMENTS}
\begin{figure*}
  \centering
  \begin{subfigure}{0.24\linewidth}
    \centering
    \includegraphics[width=\linewidth]{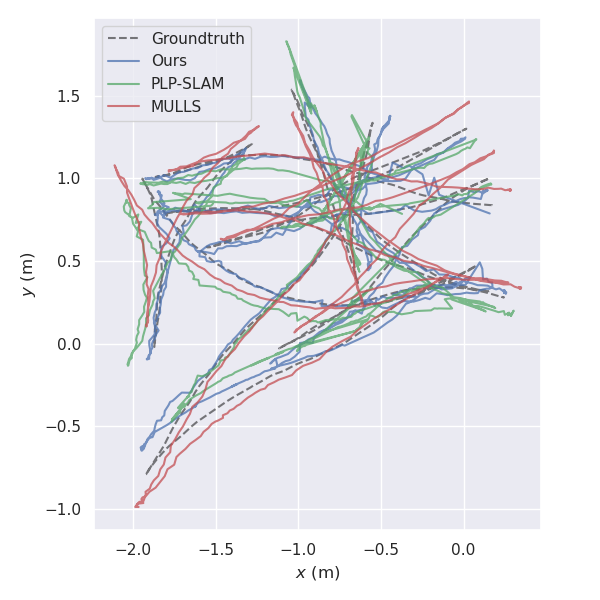}
    \caption{Sequence: room\_01}
    \label{fig:room_vs_base}
  \end{subfigure}
  \begin{subfigure}{0.24\linewidth}
    \centering
    \includegraphics[width=\linewidth]{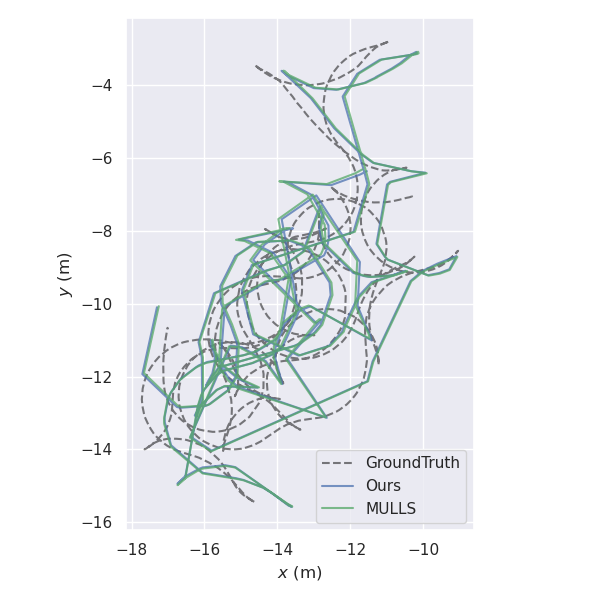}
    \caption{Sequence: hall\_03}
    \label{fig:hall_vs_base}
  \end{subfigure}
  \begin{subfigure}{0.24\linewidth}
    \centering
    \includegraphics[width=\linewidth]{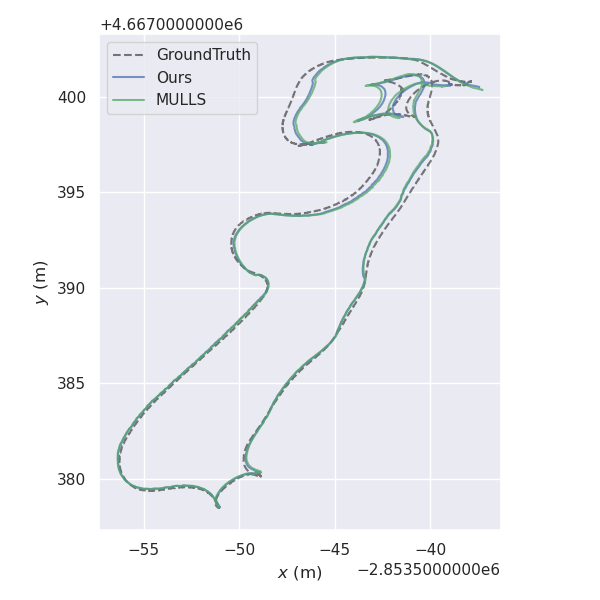}
    \caption{Sequence: gate\_01}
    \label{fig:gate_vs_base}
  \end{subfigure}
  \begin{subfigure}{0.24\linewidth}
    \centering
    \includegraphics[width=\linewidth]{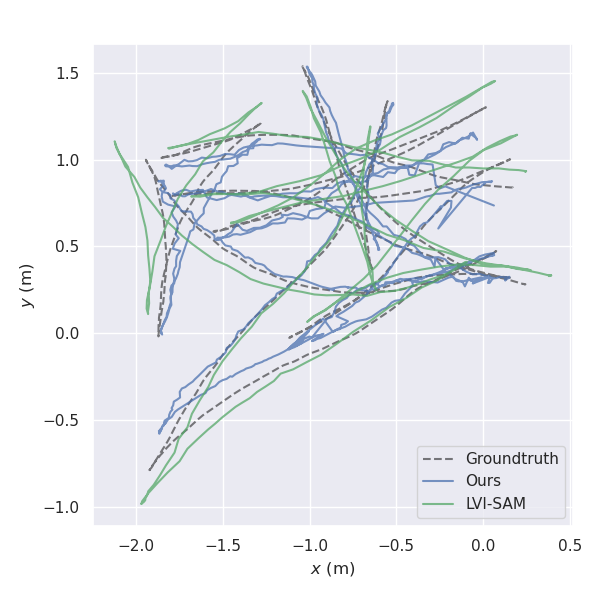}
    \caption{Sequence: room\_01}
    \label{fig:room_vs_sota}
  \end{subfigure}
\caption{The comparison regarding the trajectories 
 of the indoor/outdoor scenes. (a), (b), and (c) compare our method with Structure PLP~\cite{shu2022structure} and MULLS~\cite{pan2021mulls} while (d) compares our method with the other multi-modal SLAM, \textit{i.e.},  LVI-SAM~\cite{shan2021lvi}.}
  \vskip -3ex
\end{figure*}

\begin{table}[h]
\caption{Comparison experiments with the single-modality baselines on the M2DGR~\cite{yin2021m2dgr}, where results are presented in terms of the aligned mean translation error (m). If the method fails to initialize or track frames less than half of the total frames, we mark it ``$\times$''.}
\label{tab:table_baseline}
\begin{center}
\begin{tabular}{cccc}
\hline
Sequence & Structure PLP~\cite{shu2022structure} & MULLS~\cite{pan2021mulls} & Ours\\
\hline
room\_01 & 0.1233 & 0.1446 & \textbf{0.0848} \\
room\_02 & 0.2063 & 0.1197 & \textbf{0.0996} \\
room\_03 & 0.1143 & 0.1620 & \textbf{0.1075} \\
\hline
hall\_02 & $\times$ & 0.3612 &  \textbf{0.3517} \\
hall\_03 & $\times$ & 0.6820 &  \textbf{0.5975} \\
hall\_04 & $\times$ & 0.9049 &  \textbf{0.8886} \\
\hline
gate\_01 & $\times$ & 0.7322 & \textbf{0.6395} \\
gate\_02 & $\times$ & 0.3916 & \textbf{0.2954} \\
gate\_03 & $\times$ & 0.4165 & \textbf{0.3211} \\
\hline
\end{tabular}
\end{center}
\vskip -3ex
\end{table}

We use the M2DGR dataset~\cite{yin2021m2dgr} to evaluate the leveraged approaches. All experiments were performed on an i5-8260U laptop with $16$ GB memory. Our method is compared with diverse baselines, \textit{i.e.}, Structure PLP-SLAM~\cite{shu2022structure}, MULLS~\cite{pan2021mulls}, and other multi-modal SLAM algorithms ORB-SLAM3~\cite{campos2021orb} and LVI-SAM~\cite{shan2021lvi}. 

\subsection{Introduction of the M2DGR Dataset}
This dataset includes indoor and outdoor sequences captured using a LiDAR sensor and a monocular camera mounted on a ground robot, as shown in Figure~\ref{fig:dataset}. In indoor settings, the robot primarily operates in well-lit office environments, where dynamic objects (people) interfere with the robot's work. In larger-scale scenarios, the robot's role includes guiding visitors to specific destinations within a hall. SLAM is utilized to identify crucial landmarks and provide positional information to enhance navigation. The robot's applicability extends beyond indoor scenes and encompasses diverse outdoor domains, \textit{e.g.}, logistics operations and package transportation, where SLAM proves especially valuable for effective navigation. Furthermore, some outdoor scenarios may have limited artificial lighting conditions that do not guarantee optimal illumination. To account for this, we include nighttime outdoor sequences in this evaluation.

To explore the diversity of mobile robot applications, we select three scenarios where mobile robots are commonly employed: the texture-rich office, the spacious orderly hall, and outdoor scenes with varying lighting conditions. 

\subsection{Comparison with Single-Modal SLAMs} 

In the scenario room, each frame has numerous feature points and lines. The trajectories shown in Figure~\ref{fig:room_vs_base} demonstrate that our system successfully achieves pose estimation. Compared with Structure PLP and MULLS- in Table~\ref{tab:table_baseline}, our method delivers superior accuracy.

The hall scene is an ideal environment for geometry-based visual SLAM due to its regular structure. However, the motion blur in this scene poses a challenge to visual SLAM which has a problem with tracking enough feature points. A similar problem occurs in outdoor scenes, where distant landmarks occupy only small portions of the captured images due to the wider field of view and longer range. In addition, certain sequences in outdoor scenes are collected at night, further exacerbating the problem. In such situations, the visual SLAM system can not detect adequate feature points, leading to tracking failures. Consequently, both the Structure PLP and our visual subsystem face limitations in estimating the entire odometry in these scenarios.

LiDAR has long-range sensing capability and stability in various lighting conditions. It always provides accurate and reliable trajectory estimation. The performance of our LiDAR subsystem is shown in Table~\ref{tab:table_baseline} and in Figures~\ref{fig:hall_vs_base}, \ref{fig:gate_vs_base}. The fusion framework enhances its linear direction, increasing the accuracy of the odometry.

\begin{table}[h]
\caption{Comparison experiments with the multi-modal SLAMs LVI-SAM~\cite{shan2021lvi} and ORB-SLAM~\cite{campos2021orb} on the M2DGR dataset~\cite{yin2021m2dgr}, where results are presented in terms of the aligned mean translation error (m).}
\label{tab:table_other_method}
\begin{center}
\begin{tabular}{cccc}
\hline
Sequence& LVI-SAM \cite{shan2021lvi} & ORB-SLAM3 \cite{campos2021orb} & Ours \\
\hline
room\_01 & 0.1498 & $\times$ &  \textbf{0.0848} \\
room\_02 & 0.1176 & $\times$ & \textbf{0.0996} \\
room\_03 & 0.1588 & $\times$ & \textbf{0.1075} \\
\hline
hall\_04 & \textbf{0.8037} & $\times$  &  0.8886 \\
\hline
gate\_02 & 0.3080  & $\times$  & \textbf{0.2954} \\
gate\_03 & \textbf{0.1198} &  $\times$ & 0.3211 \\
\hline
\end{tabular}
\end{center}
\vspace{-3mm}
\end{table}

\subsection{Comparison with Multi-Modal SLAMs} 

Due to the limited availability of open-source LiDAR-visual SLAM algorithms in recent years, we choose two well-established multi-modal algorithms for comparison: the visual-inertial-LiDAR SLAM, LVI-SAM~\cite{shan2021lvi} and the visual-inertial SLAM, ORB-SLAM3~\cite{campos2021orb}.
Comparing the outdoor sequences in Table~\ref{tab:table_other_method}, LVI-SAM, which integrates data from the IMU, camera, and LiDAR sensors, harvests more accurate visual-enhanced LiDAR odometry. The visual-inertial algorithm ORB-SLAM3 suffers from significant drift and then tracking failures in all scenarios.
In spite of utilizing only two sensors, compared with previous works, our method achieves higher accuracy and robustness in both indoor and outdoor scenes, as illustrated in Figure~\ref{fig:room_vs_sota}.

\section{CONCLUSION}
In this paper, we have proposed a tightly-coupled LiDAR-visual SLAM based on geometric features, which develops a dual-system framework allowing two SLAM algorithms to run simultaneously. In the visual subsystem, we detect lines and associate depth and direction information from LiDAR geometric features to reconstruct 3D lines. Some of these reconstructed lines are employed to estimate the initial pose as new landmarks, others are used as additional parameters to constrain the line direction in the visual back-end. The LiDAR subsystem adjusts the direction of its linear features based on the complete line detected by the visual system, reducing the outlier likelihood during the registration process. These modifications further improve the accuracy of our system. In addition, we implement a detection module to detect the operation of the subsystems. If the visual subsystem fails, the LiDAR odometry is provided as system output to improve robustness.

The proposed method has demonstrated superior robustness compared to the standalone sensor frameworks, especially in scenarios involving high-speed sensor motion and significant lighting changes. Despite utilizing fewer sensors, our approach outperforms state-of-the-art methods in most cases. Our algorithm helps the mobile agent achieve localization and mapping in multiple scenarios, which enables the robot to accomplish tasks efficiently even with limited computational resources. We hope that we will obtain additional constraints between the odometry of the two subsystems to further enhance the system's performance.

\bibliographystyle{IEEEtran}
\bibliography{bib}

\end{document}